\documentclass{article}
\usepackage[nonatbib]{neurips_2022}
\nolinenumbers

\usepackage[multiple]{footmisc}
\usepackage[bookmarksnumbered,pdfpagelabels=true,plainpages=false,colorlinks=true,
            linkcolor=red,citecolor=red,urlcolor=blue]{hyperref}
            
\usepackage[utf8]{inputenc} 
\usepackage[T1]{fontenc}    
\usepackage{hyperref}       
\usepackage{booktabs}    
\usepackage{amsfonts}       
\usepackage{nicefrac}       
\usepackage{microtype}      
\usepackage{xcolor}         
\usepackage{hyperref}  
\usepackage{url}       
\usepackage{booktabs}     
\usepackage{amsfonts}   
\usepackage{amsmath}
\usepackage{nicefrac}      
\usepackage{microtype}
\usepackage{multirow}
\usepackage{wrapfig}
\usepackage{threeparttable}
\usepackage{graphicx}
\usepackage{xcolor} 
\usepackage{enumitem}
\usepackage{multicol}
\usepackage{algorithm}
\usepackage{algpseudocode}

\newcommand{\figref}[1]{Fig.~\ref{#1}}
\newcommand{\figureref}[1]{Figure~\ref{#1}}

\newcommand{\tableref}[1]{Table~\ref{#1}}

\let\oldFootnote\footnote
\newcommand\nextToken\relax

\renewcommand\footnote[1]{%
    \oldFootnote{#1}\futurelet\nextToken\isFootnote}

\newcommand\isFootnote{%
    \ifx\footnote\nextToken\textsuperscript{,}\fi}

\title{Simple Baseline for Weather Forecasting Using Spatiotemporal Context Aggregation Network}
\author{%
   Minseok Seo\thanks{Equal contribution}\\
   SI Analytics\\
   \texttt{minseok.seo@si-analytics.ai}
   \And
   Doyi Kim\footnotemark[1]\\
   SI Analytics\\
   \texttt{doyi@ewhain.net}\\
   \And
   Seungheon Shin\\
   SI Analytics\\
   \texttt{shshin@si-analytics.ai}\\
   \And
   Eunbin Kim\\
   SI Analytics\\
   \texttt{ebkim@si-analytics.ai}\\
   \And
   Sewoong Ahn\\
   SI Analytics\\
   \texttt{anse3832@si-analytics.ai}\\
   \And
   Yeji Choi\thanks{Corresponding author}\\
   SI Analytics\\
   \texttt{yejichoi@si-analytics.ai}\\
}
\begin{document}

\maketitle

\begin{abstract}
Traditional weather forecasting relies on domain expertise and computationally intensive numerical simulation systems.
Recently, with the development of a data-driven approach, weather forecasting based on deep learning has been receiving attention.
Deep learning-based weather forecasting has made stunning progress, from various backbone studies using CNN, RNN, and Transformer to training strategies using weather observations datasets with auxiliary inputs.
All of this progress has contributed to the field of weather forecasting; however, many elements and complex structures of deep learning models prevent us from reaching physical interpretations.
This paper proposes a SImple baseline with a spatiotemporal context Aggregation Network (SIANet) that achieved state-of-the-art in 4 parts of 5 benchmarks of W4C’22.
This simple but efficient structure uses only satellite images and CNNs in an end-to-end fashion without using a multi-model ensemble or fine-tuning.
This simplicity of SIANet can be used as a solid baseline that can be easily applied in weather forecasting using deep learning.
\end{abstract}
\section{Introduction}
Weather forecasting is essential to early warning and monitoring systems for natural disasters.
Timely action by accurate forecasting can mitigate the impacts.
Recently, extreme precipitation events have significantly increased in frequency and intensity over the globe, which can lead to a high potential for flooding~\cite{frnda2022ecmwf}.
Predicting rain events usually use weather radar and the Numerical Weather Prediction (NWP) model.
Over the last few decades, a very short-term forecast, nowcasting, has substantially improved in a combination of that two ways.

NWP model predicts the future weather state from the current atmospheric conditions by solving physical theories.
%
%Still, NWP models need high computational costs, and the first two hours of prediction are known to have a high prediction error.
Still, NWP models need high computational costs, and the first two hours of prediction are known to have a high prediction error because the model resolution is not enough to describe cloud particle development. 
More recently, deep learning (DL) has been attempted to combine with NWP to improve the accuracy of short-term predictions~\cite{frnda2022ecmwf, rojas2022deep}.
However, the DL models still do not fully understand the physical theories of the atmosphere, so NWP data is used just as ancillary data~\cite{espeholt2022deep, leinonen2022nowcasting} in the training process. 

The current nowcasting approach in operation is based on radar extrapolation methods.
Weather radar produces a high-resolution precipitation map, which includes the motion and intensity of rain events.
%
%Classical radar-based prediction methods extrapolate from radar maps, so the accuracy decreases as the prediction time and observation distance increase.  
Classical methods extrapolate the movement of precipitation systems from radar maps, so the accuracy of rainfall intensity decreases as the prediction time and observation distance increase.
Several data-driven DL algorithms have been used for rain forecasting to overcome these limitations (~\cite{ravuri2021skilful}). 
However, the radar can detect water droplets larger than a specific size ($\geq 2 mm$) that are matured cloud particles and may miss smaller droplets.
That means the model cannot recognize the convective initiation.

Thus, the models have been developed by considering a broader spatiotemporal context. 
The LSTM (long-short-term memory) architecture uses memory blocks that capture spatiotemporal dependencies among sequential data. 
Recently, ConvLSTM by ~\cite{shi2015convolutional} has been employed for 2D images and exploited in weather prediction models without physical theories ~\cite{espeholt2022deep, shi2017deep, ayzel2020rainnet, sonderby2020metnet}. 
~\cite{espeholt2022deep} proposed a multi-data-based prediction model, MetNet-2, which uses an axial attention module to efficiently capture the longer spatial dependencies in the data.
However, using additional static data, ConvLSTM, or temporal embedding for temporal modeling could expand complicated structures.
%
%Even in most cases, it has not been experimentally determined whether each component is significant.
%
%This complexity makes researchers suffer when doing deep learning-based weather forecasting research.

%
In this paper, we proposed a simple but efficient structure, a SImple baseline with a spatiotemporal context Aggregation Network (SIANet).
SIANet only consists of 3D-CNN compared to others that use a mixture of several models ~\cite{shi2015convolutional, gao2022earthformer}.
We use satellite image data as an input without other static data such as latitude, longitude, and topological height.
Despite this simple structure, SIANet achieved state-of-the-art in 4 out of 5 of the W4C’22 benchmark datasets.
\begin{figure*}[t!]
    \centering
    \includegraphics[width=1.0\columnwidth]{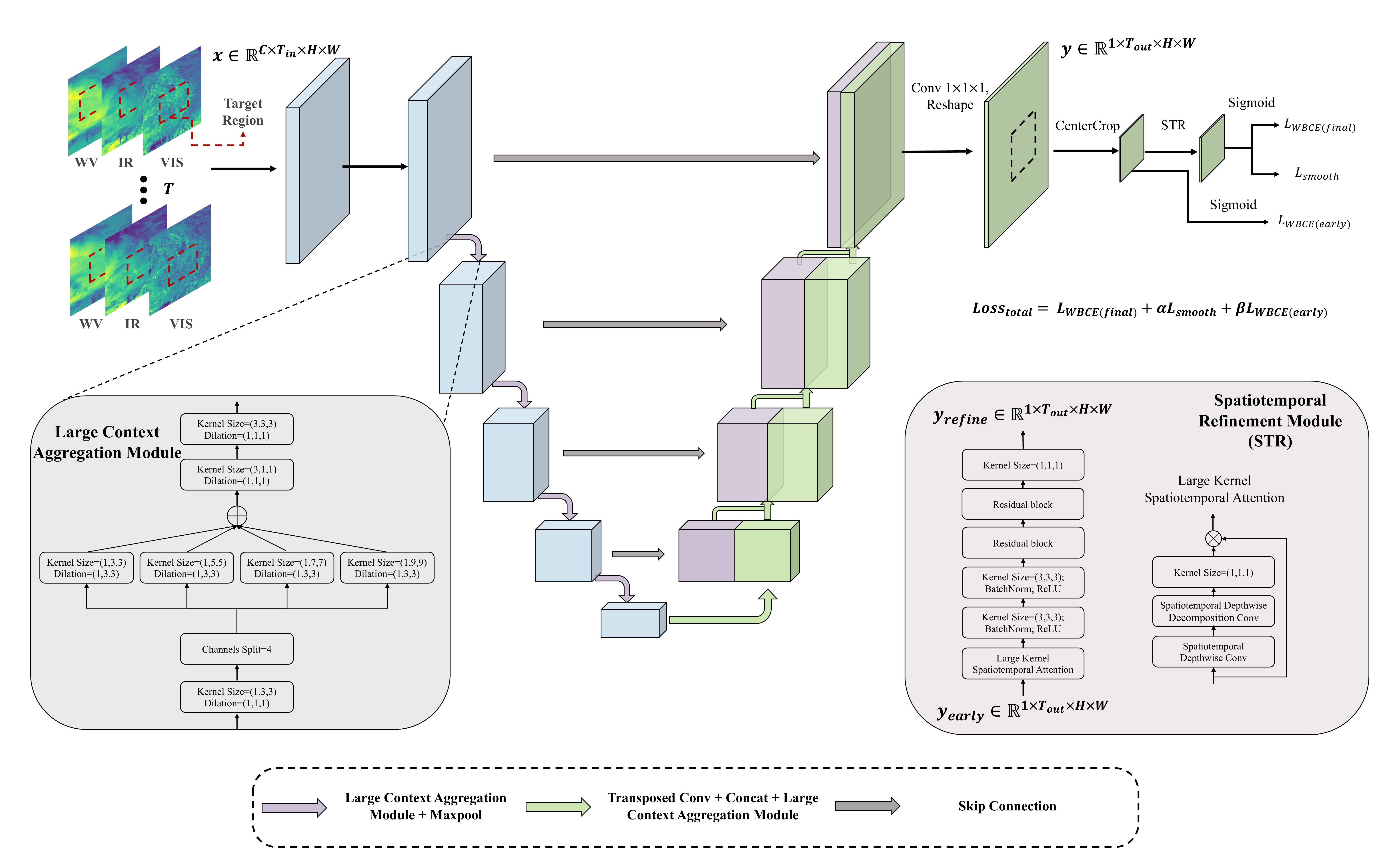}
    \caption{An overview of SIANet. SIANet is composed of pure 3D-CNN and uses a large context aggregation module as a basic block. Also, the spatiotemporal refinement module is trained end-to-end fashion.}
    \label{fig:main}
\end{figure*}
\begin{figure}[t!]
    \centering
    \includegraphics[width=0.8\columnwidth]{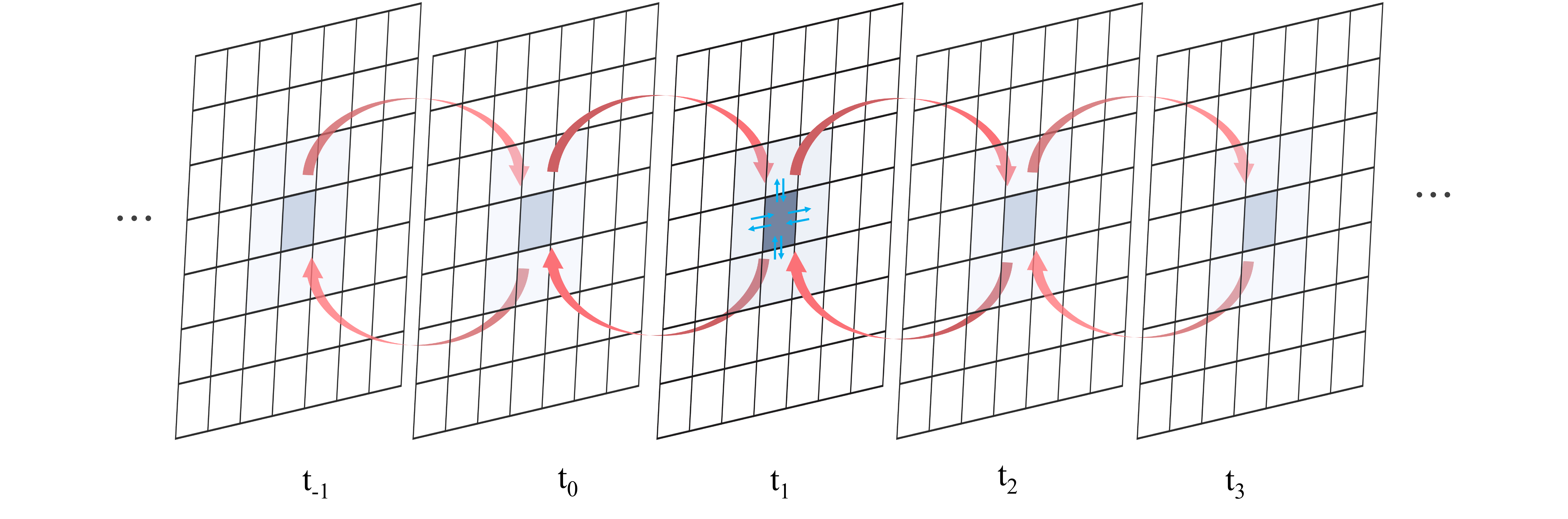}
    \caption{Motivation of the Spatiotemporal Refinement Module (STR).}
    \label{fig:STR}
\end{figure}
\section{Method}
In this section, we describe the architecture of the SIANet in detail.
SIANet predicts rainfall locations for the next 8 hours with 32-time slots from an input sequence of 4-time slots of the preceding hour.
%
%The input sequence consists of four (sequential/consecutive) 11-band spectral satellite images.
The input sequence consists of four consecutive satellite images from 11 spectral band.
These 11 channels consists of satellite radiances covering so-called visible (VIS), water vapor (WV), and infrared (IR) bands.
Each satellite image covers a 15-minute period and its pixels correspond to a spatial area of about 12 km $\times$ 12 km.
The prediction output is a sequence of 32 images representing rain rate from radar reflectivity.
Output images also have a temporal resolution of 15 minutes but have higher spatial resolution than input data, with each pixel corresponding to a spatial area of about 2 km x 2 km.
%
%So, in addition to predicting the weather in the future, converting coarse satellite inputs to fine radar outputs.
Thus, in the training process, the proposed model consider how to convert the coarse satellite images to the fine radar images for target regions in addition to predicting the weather in the future. 
\subsection{Kernel Decomposition}
If there is a 3D convolutional filter with a receptive field size of $K_{t} \times K_{h} \times K_{w}$ size, the number of Flops of the CNN is as follows:
\begin{equation}
  Flops = H \times W \times C_{in} \times C_{out} \times (K_{t} \times K_{h} \times K_{w} )
  \label{eq:overview1}
\end{equation}
where $H, W$ are the height and width of the image, $C_{in}$ is the input channel, and $C_{out}$ is the number of output channels.
If 3D convolution is separated into spatial 2D convolution and temporal 1D convolution through kernel decomposition, the number of parameters is as follows:
\begin{equation}
  Flops_{d} = H \times W \times C_{in} \times C_{out} \times (K_{t} + K_{h} \times K_{w} )
  \label{eq:overview2}
\end{equation}

\subsection{Overview}
\label{sec:over}
Weather forecasting aims to infer the future weather state based on previous conditions.
When satellite inputs $x \in \mathbb{R}^{C \times T_{in} \times H \times W}$ and ground-radar outputs $y \in \mathbb{R}^{1 \times T_{out} \times \frac{H}{6} \times \frac{W}{6}}$ are given,
SIANet $\mathcal{F}(.)$ learns to predict $y$ by inputting $x$ through loss functions $\mathcal{L}$ as shown in the following equation:
\begin{equation}
  \mathcal{L}_{bce} = -\sum_{h,w}(y\log{\mathcal{F}(x)}+(1-y)\log(1-\mathcal{F}(x))
  \label{eq:overview3}
\end{equation}
~\figureref{fig:main} shows an overview of SIANet.
The model has a simple U-Net-based structure that does not use the complex structures of recent leading approaches such as RNN, CNN+RNN, transformer series, or additional information such as time embedding, latitude, longitude, and topological height.
SIANet uses the Large Spatial Context Aggregation Module (LCAM) as a basic block.
It is an architecture design based on matrix decomposition~\cite{geng2021attention, guo2022segnext}, and the amount of calculation is small compared to parameters.
The final component of SIANet is the spatiotemporal refinement module, which is trained end-to-end without additional training.
It was designed with inspiration from the Markov chain that post-processes through spatio-temporal correlation in the conventional weather forecasting field.

\subsection{Large Spatial Context Aggregation Module}
\label{sec:scam}
The work most similar to the LCAM is a Visual Attention Network (VAN)~\cite{guo2022visual} that recently achieved higher performance than the transformer architecture with decomposition convolution.
LCAM uses decomposition convolutions similar to VAN, but uses Multi Kernel Split Convolutions (MKSC) instead of depthwise convolutions.

As depicted in ~\figref{fig:main}, LCAM contains three parts: a multi-kernel split convolutions to capture multi receptive field, spatiotemporal decomposition convolutions to capture spatiotemporal decomposed information, and an $3 \times 3 \times 3$ convolution to aggregate spatiotemporal information.
When a feature map $F\in \mathbb{R}^{C \times T \times H \times W}$ is given, LCAM is expressed by the following algorithm:
\begin{algorithm}
    \caption{LCAM}\label{euclid}
    \begin{algorithmic}
        \State \textbf{Input:} $F \in \mathbb{R}^{C \times T \times H \times W}$
        \State $F = SpatioConv_{1 \times 3 \times 3}(F)$
        \State $F_{1}, F_{2}, F_{3}, F_{4} = ChannelsSplit(F)$ \Comment{Divide the channels at even intervals}
        \For {$i$ in $4$}
            \State $\hat{F_{i}} = DilateConv_{1 \times n[i] \times n[i]}(F_{i})$ \Comment{$n = [3, 5, 7, 9]$}
        \EndFor
        \State $\hat{F} = Concat(\hat{F_{1}}, \hat{F_{2}}, \hat{F_{3}}, \hat{F_{4}})$ 
        \State $\hat{F} = TemporalConv_{3 \times 1 \times 1}(\hat{F})$
        \State $\hat{F} = Conv_{3 \times 3 \times 3}(\hat{F})$ \Comment{Spatiotemporal aggregation}        
    \end{algorithmic}
    \label{alg:LCAM}
\end{algorithm}
Note that the size of the dividing channels is a hyperparameter.
In the W4C’22 dataset, the highest performance was achieved when the channel division size was set to 2 (3x3,5x5).
However, we empirically found that dividing the channel size by 4 performed best on video prediction dataset~\cite{srivastava2015unsupervised}.

\subsection{Spatiotemporal Refinement Module}
\label{sec:STR}
Conditional random field (CRF) has been widely used as a post-processing algorithm in object semantic segmentation and has shown robust results against various noises.
Recently, a deep learning-based post-processing method has also been proposed and has shown promising results~\cite{huynh2021progressive}.
The accuracy of weather forecasting is significantly affected by the post-processing method.
In particular, the post-processing algorithm is essential because weather is highly spatially and temporally related.

\begin{figure}[!ht]
    \centering
    \includegraphics[width=0.6\columnwidth]{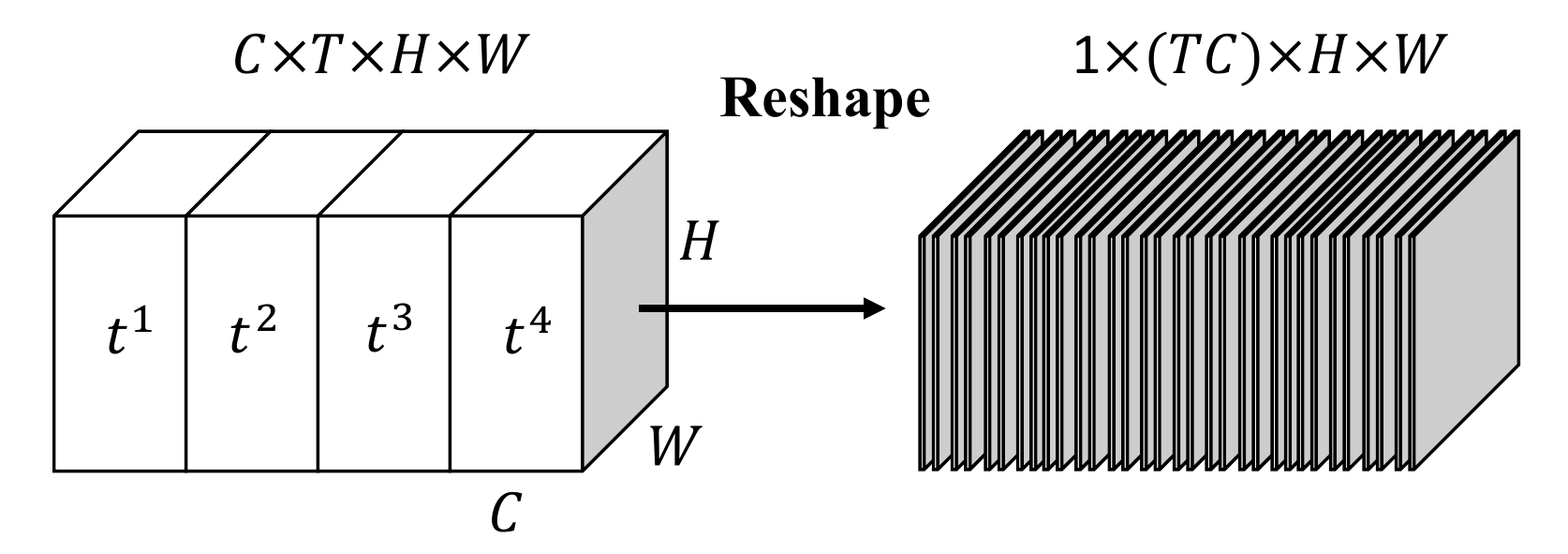}
    \caption{The reshape method adopted by SIANet.}
    \label{fig:reshape}
\end{figure}

We propose the Spatiotemporal Refinement Module (STR) to solve this problem.
~\figureref{fig:STR} shows our motivation. 
As shown in the figure, since adjacent pixels are correlated with each other, using the spatiotemporal modeling can improve the performance of the weather forecasting model.
The STR is trained for the purpose of refinement of the output $y_{early} \in \mathbb{R}^{1 \times T \times H \times W}$ from the 3D-UNet structure based on the LCAM block. $y_{early}$ is generated through reshaping after channel reduction with target region crop and $1 \times 1 \times1$ convolution.
%
%~\figureref{fig:reshape} shows the process by which the channels and the time dimensions are multiplied to create the target time dimension that SIANet should use for weather forecasting.
~\figureref{fig:reshape} shows the process how the channels and the time dimensions are multiplied to create the target time dimension used for weather forecasting.

As shown in ~\figref{fig:main}, the STR consists of a large kernel spatiotemporal attention and two sequential residual blocks.
The large kernel spatiotemporal attention re-weights $y_{early}$ by considering the receptive field as much as $7\times7\times7$ on the spatiotemporal axis of the softmax value of each pixel of $y_{early}$.
Note that large kernel spatiotemporal attention is a temporally extended version of VAN~\cite{guo2022visual}.
After that, $y_{final} \in \mathbb{R}^{1 \times T \times H \times W}$, the result of refinement of $y_{early}$, is output by aggregating the surrounding spatiotemporal information through the residual block with $3\times 3\times 3$ kernel.
The loss function of SIANet including STR follows the equation below:
\begin{equation}
  \mathcal{L}_{total} = \mathcal{L}_{bce}(y_{final}, y) + \alpha\mathcal{L}_{bce}(y_{early}, y)
  \label{eq:STR}
\end{equation}
where $\alpha$ is the weighting factor and is set to 0.2 in all our experiments.

\section{Experiments}
We used the \textit{W4C'22} stage1 benchmark dataset and the \textit{W4C'22} stage2 benchmark dataset to evaluate SIANet.
Also, our SIANet was evaluated on the \textit{W4C'22} core transfer benchmark dataset and achieved stage-of-the-art on the test set and held-out set. However, these results are not covered in this paper, but in our core transfer paper.

In this section, the experimental results of the W4C’22 leaderboard and the ablation study experimental results of LCAM and Refine module are described in detail.
The performance evaluation metric of all experiments was selected as mean intersection over union (mIoU).
Note that all ablation studies were performed on the stage 1 testset.
The code is available on \footnote{\tiny	{https://github.com/seominseok0429/W4C22-Simple-Baseline-for-Weather-Forecasting-Using-Spatiotemporal-Context-Aggregation-Network}}
\subsection{Experimental Setting}
\paragraph{\textit{W4C’22 stage1} Datasets} consist of three European regions
  selected based on precipitation characteristics.
The task of the dataset is to receive four satellite images at 15-minute intervals and predict 32 rain events at 15-minute intervals at the pixel level.
Therefore, it is a binary classification task that predicts rain/no-rain at the pixel-level.
OPERA data was separated into rain/no-rain pixels by a threshold of 0.001 mm/hr on ground radar, and the stage1 dataset covered February to December 2019.
\paragraph{\textit{W4C’22 stage2} Datasets} consists of two years, 2019 and 2020, and seven regions of Europe (R15, R34, R76, R04, R05, R06, R07).
Unlike the stage1 dataset, the rain/no-rain threshold is 0.2 mm/hr, and latitude, longitude, and topological height are provided.
Note that in all our experiments, only satellite images are used among them.
\paragraph{Implementation detail} 
To train SIANet, the batch size of one GPU was set to 16, and FP16 training was used.
In addition, 90 epochs and 4 positive weights of binary cross entropy were used.
The initial learning rate was set to 1e-4, weight decay to 0.1, and dropout rate to 0.4.
AdamW was used as the optimizer, and the learning rate was reduced by 0.9 when the loss was higher than the validation loss at the previous epoch.
All experiments were performed on Nvidia A100 $\times$ 8 GPUs.
\begin{table*}[!ht]
\centering
\renewcommand{\arraystretch}{1.5}
\caption{\textit{W4C’22 Stage1 test leaderboard} experiment results. Note that bold indicates the highest performance.}
\label{tab:tab1}
\resizebox{0.9\columnwidth}{!}{%
\begin{tabular}{l|cccccccccccl}
\hline
\hline
\multicolumn{1}{c|}{}       & \multicolumn{12}{c}{\textbf{\textit{W4C'22 Stage 1 Test Leaderboard}}}                                                                                                             \\ \cline{2-13} 
\multicolumn{1}{c|}{\textbf{Method}} & \multicolumn{4}{c|}{\textbf{Boxi15}}                        & \multicolumn{4}{c|}{\textbf{Boxi34}}                        & \multicolumn{4}{c}{\textbf{Boxi76}}    \\ \cline{2-13} 
\multicolumn{1}{c|}{}       & Precision & Recall & F1 & \multicolumn{1}{c|}{IoU} & Precision & Recall & F1 & \multicolumn{1}{c|}{IoU} & Precision & Recall & F1 & IoU \\ \hline
SIANet & 0.617 & \textbf{0.751} & \textbf{0.678}& \multicolumn{1}{c|}{\textbf{0.512}} &  0.535&     \textbf{0.760}   &  \textbf{0.628}  & \multicolumn{1}{c|}{\textbf{0.458}} &      0.649    &    0.550    &   \textbf{0.595} &  \textbf{0.424}   \\
FIT-CTU & 0.629 & 0.723 & 0.673& \multicolumn{1}{c|}{0.507} &  0.555&     0.705   &  0.621  & \multicolumn{1}{c|}{0.450} &      0.653    &    0.489    &   0.559 &  0.388   \\
MS-nowcasting & 0.647 & 0.670 & 0.658& \multicolumn{1}{c|}{0.490} &  0.537&     0.665   &  0.594  & \multicolumn{1}{c|}{0.423} &      0.554    &    0.572    &   0.563 &  0.392   \\
meteoai & 0.647 & 0.669 & 0.658& \multicolumn{1}{c|}{0.490} &  0.517&     0.684   &  0.589  & \multicolumn{1}{c|}{0.417} & 0.551    &    0.576    &   0.563 &  0.392   \\
KAIST AI & 0.587 & 0.707 & 0.641& \multicolumn{1}{c|}{0.472} &  0.506&     0.749   &  0.604  & \multicolumn{1}{c|}{0.433} &      0.556    &    0.575    &    0.565 &   0.394   \\
antfugue & 0.577 & 0.707 & 0.636& \multicolumn{1}{c|}{0.466} &  0.547&      0.706   &  0.616  & \multicolumn{1}{c|}{0.446} &      0.665    &    0.479    &   0.557 &  0.386   \\
 \hline \hline
\end{tabular}%
}
\end{table*}
\subsection{Stage1 Results}
\paragraph{Results} ~\tableref{tab:tab1} is the result of evaluating SIANet on the stage1 test set.
As shown in the table, our SIANet achieved state-of-the-art performance in all regions of R15, R34, and R76.
We achieved these experimental results without additional deep learning model training processes such as multi-model ensemble and fine-tuning, or complex model structures such as ConvLSTM, hybrid, and lead time embedding.
%
%These experimental results indicate that SIANet is simple but efficient.
%
\paragraph{Discussion} SIANet is an end-to-end model composed of only 3D-CNN.
In this paper, we did not use tricks such as fine-tuning and multi-model ensemble to experimentally prove that SIANet has a simple structure but a strong performance.
However, we believe that SIANet will also benefit from additional performance improvements by applying a fine-tuning, multi-model ensemble.

\begin{table}[!ht]
\centering
\renewcommand{\arraystretch}{1.5}
\caption{\textit{W4C’22 Stage2 test leaderboard} experiment results. Note that bold indicates the highest
performance.}
\label{tab:tab2}
\resizebox{0.9\columnwidth}{!}{%
\begin{tabular}{c|cccllllcccllll|c}
\hline
\hline
\multicolumn{1}{l|}{} & \multicolumn{14}{c|}{W4C 22 Test Leaderboard}                                                                                      &       \\ \cline{2-15}
Method                & \multicolumn{7}{c|}{2019}                                                  & \multicolumn{7}{c|}{2020}                             & mIoU  \\ \cline{2-15}
\multicolumn{1}{l|}{} & R15   & R34   & R76   & R04   & R05   & R06   & \multicolumn{1}{l|}{R07}   & R15   & R34   & R76   & R04   & R05   & R06   & R07   &       \\ \hline
SIANet                & 0.305 & 0.177 & \textbf{0.360} & \textbf{0.260} & \textbf{0.317} & \textbf{0.422} & \multicolumn{1}{l|}{0.369} & 0.270 & \textbf{0.392} & 0.128 & \textbf{0.361} & \textbf{0.277} & \textbf{0.345} & \textbf{0.244} & \textbf{0.302} \\
meteoai               & \textbf{0.310} & 0.180 & 0.323 & 0.244 & 0.309 & 0.418 & \multicolumn{1}{l|}{\textbf{0.375}} & 0.220 & 0.384 & \textbf{0.163} & 0.343 & 0.269 & 0.285 & 0.207 & 0.288 \\
FIT-CTU               & 0.308 & 0.194 & 0.350 & 0.253 & 0.308 & 0.411 & \multicolumn{1}{l|}{0.333} & \textbf{0.276} & 0.388 & 0.106 & 0.297 & 0.249 & 0.283 & 0.199 & 0.283 \\
KAIST-AI              & 0.269 & \textbf{0.200} & 0.293 & 0.233 & 0.280 & 0.383 & \multicolumn{1}{l|}{0.361} & 0.237 & 0.362 & 0.104 & 0.320 & 0.271 & 0.335 & 0.133 & 0.270 \\ \hline\hline
\end{tabular}%
}
\end{table}
\begin{table}[!ht]
\centering
\renewcommand{\arraystretch}{1.5}
\caption{\textit{W4C’22 Stage2 Heldout leaderboard} experiment results. Note that bold indicates the highest
performance.}
\label{tab:tab3}
\resizebox{0.9\columnwidth}{!}{%
\begin{tabular}{c|cccccccccccccc|c}
\hline \hline
          & \multicolumn{14}{c|}{W4C 22 Heldout Leaderboard}                                                                                      &       \\ \cline{2-15}
Method    & \multicolumn{7}{c|}{2019}                                                  & \multicolumn{7}{c|}{2020}                             & mIoU  \\ \cline{2-15}
          & R15   & R34   & R76   & R04   & R05   & R06   & \multicolumn{1}{c|}{R07}   & R15   & R34   & R76   & R04   & R05   & R06   & R07   &       \\ \hline
FIT-CTU   & \textbf{0.382} & 0.290 & \textbf{0.209} & 0.310 & \textbf{0.365} & 0.419 & \multicolumn{1}{c|}{0.244} & 0.268 & \textbf{0.316} & 0.420 & \textbf{0.360} & \textbf{0.358} & 0.436 & 0.043 & \textbf{0.316} \\
meteoai   & 0.334 & 0.295 & 0.198 & 0.313 & 0.318 & 0.410 & \multicolumn{1}{c|}{\textbf{0.287}} & 0.259 & 0.260 & \textbf{0.448} & 0.348 & 0.343 & 0.439 & \textbf{0.047} & 0.307 \\
SIANet    & 0.343 & \textbf{0.300} & 0.206 & \textbf{0.320} & 0.350 & \textbf{0.434} & \multicolumn{1}{c|}{0.216} & 0.249 & 0.280 & 0.417 & 0.350 & 0.328 & 0.446 & 0.020 & 0.304 \\
team-name & 0.321 & 0.278 & 0.195 & 0.305 & 0.341 & 0.395 & \multicolumn{1}{c|}{0.236} & \textbf{0.270} & 0.306 & 0.402 & 0.330 & 0.339 & \textbf{0.470} & 0.001 & 0.299 \\ \hline \hline
\end{tabular}%
}
\end{table}
\subsection{Stage2 Results}
\paragraph{Test Leaderboard Results}  ~\tableref{tab:tab2} is the result of evaluating SIANet on the stage2 test set.
SIANet achieved stage-of-the-art performance in regions R34, R76, R05, R06, and R07 in 2019 and regions R34, R04, R05, R06, and R07 in 2020.
In this experiment, SIANet did not use the longitude, latitude, and topological height information provided by the W4C'22 stage2 dataset, but only satellite images.
These experimental results indicate that SIANet can achieve high performance using only satellite images and a simple model structure.
\paragraph{Heldout Leaderboard Results}  ~\tableref{tab:tab3} is the result of evaluating SIANet on the stage2 heldout set.
SIANet achieved third place performance with 0.304 mIoU performance.
This result is 0.012 lower than the 0.316 scores of the first place.
In addition, SIANet scored $0.302 \sim 0.309$ in the test set, validation set, and held-out set.
These experimental results indicate that SIANet is not a model whose performance varies greatly depending on the test set, but a solid baseline with consistent performance in various test sets.
\paragraph{Discussion} Longitude, latitude, and topological height are important properties that can reflect regional features in the model.
Many deep learning-based weather forecasting works have improved performance by utilizing static information.
%
%SIANet can also concatenate and input the static information to the input image.
%
In this paper, we did not experiment using static data in SIANet for simplicity, but we believe that SIANet can easily add static data because it is easy to modify and has a simple structure, and can improve performance.
\subsection{Ablation Study}
\begin{table}[!ht]
\renewcommand{\arraystretch}{1.2}
\centering
\caption{Efficiency comparison experiment between SIANet and UNet3D according to init filter size}
\label{tab:efficay}
\begin{tabular}{ccccc}
\hline
\hline
\textbf{Models}  & \textbf{Init Filter Size} & \textbf{FLOPs} $\downarrow$  & \textbf{Parameters} $\downarrow$ & \textbf{mIoU} $\uparrow$ \\ \hline
U-Net3D & 32               & \textbf{39.52}  & 5.7M       & 0.332 \\
SIANet  & 32               & 54.30  & \textbf{4.6M}       & \textbf{0.447} \\ \hline
U-Net3D & 64               & 152.22 & 22.6M      & 0.338 \\
SIANet  & 64               & \textbf{137.76} & \textbf{17.9M}      & \textbf{0.465} \\ \hline \hline
\end{tabular}
\end{table}
\paragraph{Efficiency Experiment} ~\tableref{tab:efficay} is an experiment table comparing the efficiency of UNet3D and SIANet according to the init filter size.
As shown in the table, SIANet not only has higher performance, but also has fewer parameters and FLOPs than UNet3D.
These experimental results are because SIANet applied decomposition method in all parts of spatial, temporal, and channels (split).
\paragraph{Efficiency Discussion} Since satellite data is accumulated on a daily basis, the quantity is very large.
Therefore, a very large-scale GPU environment is essential to use all weather data for training.
Recently proposed transformer-based weather prediction models show promising performance, but FLOPs are very large. (Transformer-based methods have large FLOPs even when parameters are small.)
However, methods requiring large FLOPs require more GPU environments to train large amounts of weather data.
The matrix decomposition technology is a method that can achieve high performance while reducing the amount of computation.
Also, SIANet experimentally showed that the matrix decomposition technique is effective for weather data.
Therefore, we hope that the matrix decomposition technique will be widely used in the field of weather, which requires training on large data.
\paragraph{Component Ablation Study} ~\tableref{tab:able} is the result of the ablation study for each component of SIANet.
The thresholds in the table are experimental results adjusted from 0.5 to 0.6, and different thresholds for each evaluation area were not used.
Experimentally, using a different threshold for each region improved performance slightly, but it was not used because of the possibility of overfitting to the testset.
In addition, CenterCrop, STM, and LCAM all showed performance improvements, and these experimental results indicate that each component has a complementary relationship with each other. Strategies are the training strategies of domain generalization~\cite{seo2022domain}.
\begin{figure}[t!]
    \centering
    \includegraphics[width=0.7\columnwidth]{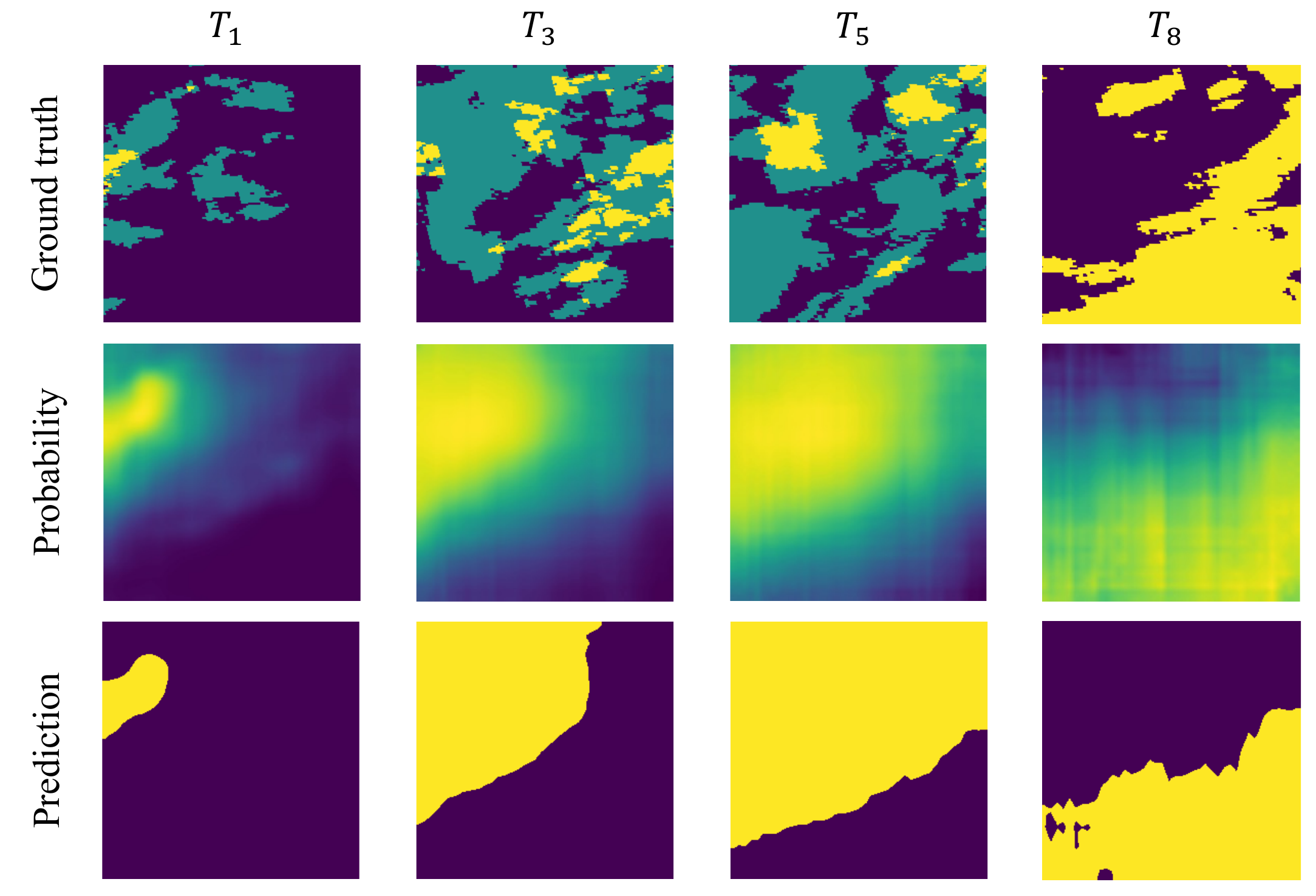}
    \caption{Predictions on the validation set using SIANet. The green area of ground trues has a rain threshold of 0, and the yellow area has a rain threshold of 0.2. SIANet's prediction result probability value for each pixel is closer to 1 as it is closer to yellow.}
    \label{fig:vis1}
\end{figure}
\begin{table}[!ht]
\renewcommand{\arraystretch}{1.2}
\centering
\caption{Results of ablation study of each component of SIANet. Note that init filter size is 64}
\label{tab:able}
\resizebox{0.9\columnwidth}{!}{%
\begin{tabular}{c|ccccc|c|l}
\hline
\hline
\textbf{Models}  & \multicolumn{5}{c|}{\textbf{Componets}}                                                                                    & \textbf{mIoU}                       & \textbf{Gain} \\ \cline{2-6}
        & \textbf{Threshold}            & \textbf{CenterCrop}           & \textbf{STM}                  & \textbf{LCAM}                 & \textbf{Strategies}            &                            &      \\ \hline
SIANet (baseline) &          -            &     -                 & -                     & -                     & -                      & 0.338                      &    +0  \\
SIANet  &           \checkmark           &           -           &        -              &        -              &    -                   & 0.359                      &     +2.1 \\
SIANet  &            \checkmark          &       \checkmark               &          -            &     -                 &     -                  & 0.381                      & +4.3     \\
SIANet  &                \checkmark      &       \checkmark               &    \checkmark                  &     -                 &                 -      & 0.418                      &    +8.0  \\
SIANet  & \multicolumn{1}{c}{\checkmark} & \multicolumn{1}{c}{\checkmark} & \multicolumn{1}{c}{\checkmark} & \multicolumn{1}{c}{\checkmark} & \multicolumn{1}{c|}{-} & \multicolumn{1}{c|}{0.443} & +10.5     \\
SIANet  & \multicolumn{1}{c}{\checkmark} & \multicolumn{1}{c}{\checkmark} & \multicolumn{1}{c}{\checkmark} & \multicolumn{1}{c}{\checkmark} & \multicolumn{1}{c|}{\checkmark} & \multicolumn{1}{c|}{\textbf{0.465}} & +\textbf{12.7}     \\ \hline \hline
\end{tabular}}
\end{table}
\subsection{Qualitative results} 
\begin{figure}[t!]
    \centering
    \includegraphics[width=0.7\columnwidth]{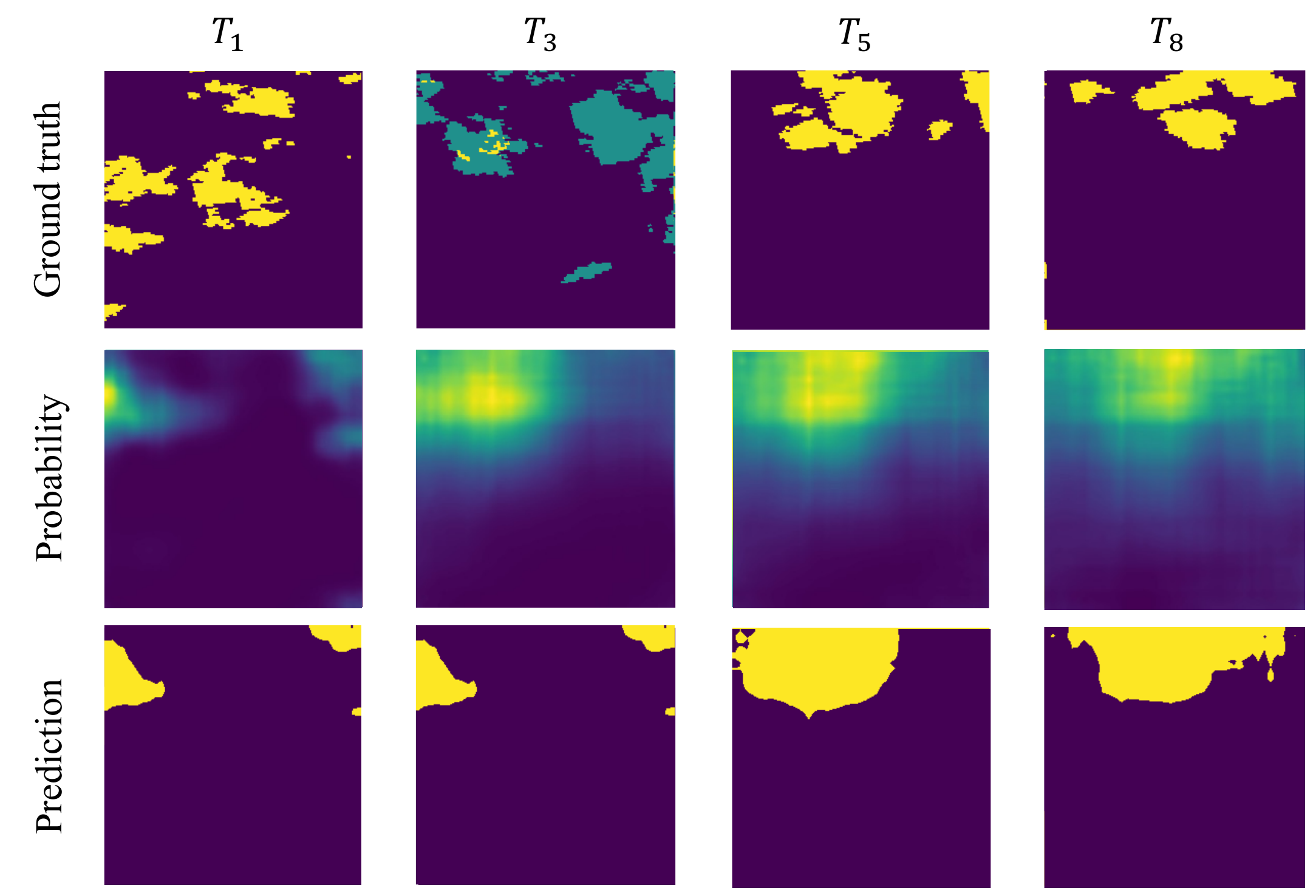}
    \caption{An example of SIANet's ability to capture the moving direction of clouds.}
    \label{fig:vis2}
\end{figure}

\paragraph{Results} ~\figureref{fig:vis1} and ~\figureref{fig:vis2} are the prediction results of SIANet. $T_{1}$ represents +1 hour, $T_{3}$ represents +3, $T_{5}$ represents +5, and $T_{8}$ represents +8 hours. As shown in ~\figref{fig:vis1}, it can be seen that SIANet makes very different predictions for each timestamp. These experimental results indicate that SIANet's $T_{1}$ to $T_{8}$ prediction values make different predictions for each timestamp, rather than predicting one average value.

~\figureref{fig:vis2} is an example of SIANet's ability to capture the direction of cloud movement.
As shown in the figure, it was confirmed that the prediction of SIANet changes depending on the moving direction of rain clouds from left to right by time period.
These experimental results indicate that SIANet made predictions by capturing the moving direction of clouds.
\paragraph{Discussion} According to the qualitative experimental results, SIANet predicts quite well when the rain rate threshold is zero.
However, when the rain rate was 0.2, there was a problem of predicting that too many areas would rain.
These experimental results are because SIANet was trained in the same rain class when the rain rate was over 0.2 without considering the rain rate.
In other words, since it has not been trained by classifying rain rates (e.g. light rain, medium rain, strong rain), SIANet has no ability to classify them.
In our future work, we plan to improve the performance by adding rain rate estimation loss to SIANet, allowing SIANet to discriminate between light, medium and strong rain.
%
%\subsection{Ablation Study}
%

\section{Conclusions}
In this paper, we proposed SIANet, a simple but efficient architecture.
We point out that existing weather forecasting models have increasingly complicated training processes, such as multi-model ensemble, fine-tuning, and auxiliary inputs.
We proposed a simple but efficient structure requiring fewer parameters, Flops, and input data, SIANet.
In addition, we propose a large spatial context aggregation module as a basic block of SIANet, composed of decomposition technology, a recent learning approach, and pure 3D-CNN.
Finally, we proposed a novel refinement module inspired by the \textit{Markov chain}.
We show that SIANet can achieve state-of-the-art on various benchmark test sets using only satellite images and without multi-model ensemble and fine-tuning.
We believe that SIANet will become a solid baseline that solves the problem that existing deep learning-based weather forecasting models are not utilized due to their complexity.

{\small
\bibliographystyle{plain}
\bibliography{egbib}
}

\end{document}